\documentclass{article}

\PassOptionsToPackage{numbers, compress}{natbib}
\usepackage[preprint]{neurips_2023}

\usepackage[utf8]{inputenc} 
\usepackage[T1]{fontenc}    
\usepackage{hyperref}       
\usepackage{url}            
\usepackage{booktabs}       
\usepackage{amsfonts}       
\usepackage{nicefrac}       
\usepackage{microtype}      
\usepackage{xcolor}         

\usepackage{bm}
\usepackage{amsmath}
\usepackage{amssymb}
\usepackage{mathtools}
\usepackage[pdftex]{graphicx}

\usepackage{subcaption}

\DeclareMathOperator*{\argmin}{arg\,min}

\newcommand\restr[2]{{
  \left.\kern-\nulldelimiterspace 
  #1
  \vphantom{|}
  \right|_{#2}
}}

\title{Short and Straight: \\Geodesics on Differentiable Manifolds}

\author{
  Daniel Kelshaw \\
  Imperial College London \\
  \texttt{djk21@imperial.ac.uk} \\
  \And%
  Luca Magri \\
  Imperial College London, \\
  Alan Turing Institute \\
  \texttt{l.magri@imperial.ac.uk} \\
}

\begin{document}

\maketitle

\begin{abstract}
  Manifolds discovered by machine learning models provide a compact representation of the underlying data. Geodesics on
  these manifolds define locally length-minimising curves and provide a notion of distance, which are key for reduced-order
  modelling, statistical inference, and interpolation. In this work, we first analyse existing methods for computing 
  length-minimising geodesics. We find that these are not suitable for obtaining valid paths, and thus, geodesic distances.
  We remedy these shortcomings by leveraging numerical tools from differential geometry, which provide the means to obtain
  Hamiltonian-conserving geodesics. Second, we propose a model-based parameterisation for distance fields and geodesic 
  flows on continuous manifolds. Our approach exploits a manifold-aware extension to the Eikonal equation, eliminating
  the need for approximations or discretisation. Finally, we develop a curvature-based training mechanism, sampling and
  scaling points in regions of the manifold exhibiting larger values of the Ricci scalar. This sampling and scaling 
  approach ensures that we capture regions of the manifold subject to higher degrees of geodesic deviation. Our proposed
  methods provide principled means to compute valid geodesics and geodesic distances on manifolds. This work opens 
  opportunities for latent-space interpolation, optimal control, and distance computation on differentiable manifolds.
\end{abstract}



\section{Introduction} \label{sec:introduction}
The manifold hypothesis states that many high-dimensional datasets are expressable on embedded low-dimensional 
manifolds~\citep{fefferman2016TestingManifoldHypothesis}. Machine learning models which utilise embeddings are well equipped
to discover these low-dimensional manifolds and provide compact representation of the underlying data. One notable example
of this is the use of autoencoders. It it well-known that an autoencoder with purely linear activations is identical in 
form to conducting principal component analysis. Introducing nonlinearities increases the expressive power of the network, 
allowing for a more compact representation~\citep{hinton2006ReducingDimensionalityData, hinton1993AutoencodersMinimumDescription}.
The latent space of the autoencoder provides a set of intrinsic coordinates on the resulting manifold~\citep{magri2022InterpretabilityProperLatent}.

The composition of transformations defining a model induce a metric which allows for principled operation on the 
underlying manifold~\citep{lee2018IntroductionRiemannianManifolds, carmo2018DifferentialGeometryCurves}. We can leverage
information about the metric to define trajectories on differentiable manifolds that provide some notion of energy-conservation,
or length-minimising paths: known as geodesics. Operating on the manifold has numerous consequences, not least: 
latent-space interpolation for generative modelling~\citep{arvanitidis2021LatentSpaceOddity, bojanowski2019OptimizingLatentSpace}; 
identifying physically-motivated manifolds such as attractors of dynamical systems\citep{page2021RevealingStateSpace, racca2022ModellingSpatiotemporalTurbulent}; 
measures of similarity and distance~\citep{kim2007DistancePreservingDimension, crane2013GeodesicsHeatNew, chen2018MetricsDeepGenerative};
as well as the ability to compute statistics on the manifold~\citep{bacak2014ComputingMediansMeans, pennec2006IntrinsicStatisticsRiemannian}.

In §\ref{sec:review_on_differential_geometry} we provide a concise review on differential geometry with a particular
focus on Riemannian manifolds and introducing the notion of geodesics. Next, we discuss principled numerical methods to 
obtain valid geodesics in §\ref{sec:valid_geodesics} and introduce a commonly adopted machine learning method. We apply
these methods to find trajectories in the latent space defined by an autoencoder and comment on the validity of the
approaches. Finally, we provide a method for computing distance fields, and geodesic flows on differentiable manifolds 
in §\ref{sec:eikonal_equation}, based on manifold-informed extensions to the Eikonal equation.

\section{Review on Differential Geometry} \label{sec:review_on_differential_geometry}
Operating on Euclidean spaces is a familiar concept with definitions for distances and angles stemming directly from
the Euclidean norm, defined by the standard inner-product. In transitioning to generic differentiable manifolds, these
familiar definitions no longer hold and a generalised mathematical framework is required. In this section we first
introduce the concept of Riemannian manifolds, endowed with a metric to enable computation of the inner product on the
tangent space. We then leverage this metric to provide a notion of directional derivatives and conditions required to 
transport a vector along the manifold in such a manner that parallelism is maintained. We take the idea of parallel
transport and define geodesics, the generalisation of straight lines on the manifold. Finally, we comment on the
computational framework employed throughout this paper, utilising automatic differentiation to compute quantities of
interest. We refer the reader to~\citet{lee2018IntroductionRiemannianManifolds, carmo2018DifferentialGeometryCurves} for
a more extensive overview of differential geometry where necessary.

\subsection{Riemannian Manifolds}
A Riemannian manifold~\citep{lee2018IntroductionRiemannianManifolds} is a pair $(M, g)$, where $M$ is a smooth manifold, 
and $g$ is a choice of Riemannian metric on $M$. This metric allows us to compute a smooth $(0, 2)$-tensor field whose 
value $g_p$ at a point $p \in M$ is an inner product on the tangent space, $T_p M$. Components of the metric are defined
as $g_{ij}(p) = \langle \restr{\partial_i}{p}, \restr{\partial_j}{p} \rangle$, where $\partial_i = \nicefrac{\partial}{\partial x^i}$
denotes the $i$th coordinate vector field. For arbitrary vectors $v, w \in T_p M$, we can compute the inner product

\begin{equation}
  \langle v, w \rangle_g = \langle v^i \restr{\partial_i}{p}, w^j \restr{\partial_j}{p} \rangle = g_p(v, w) = g_{ij} v^i w^i,
\end{equation}

providing the ability to compute distances and angles on the tangent-plane. Note that here, and throughout this paper, 
we use the Einstein summation convention in conjunction with notation $x^i, x_i$ to denote contravariant and covariant
vectors accordingly. To provide a connection to Euclidean space, we consider the Euclidean metric $\tilde{g}_{ij} = \delta_{ij}$; 
this allows us to recover our standard definition of the inner product for Euclidean spaces, $\langle v, w \rangle_{\tilde{g}} = v^i w^i$.
An important point of note is that every smooth manifold admits a Riemannian metric.

We next consider the case of embedded submanifolds. Suppose $(\tilde{M}, \tilde{g})$ is a Riemannian manifold and 
$M \subseteq \tilde{M}$ is an embedded submanifold. Given a smooth immersion $\iota: M \hookrightarrow \tilde{M}$, the 
metric $g = \iota^\ast \tilde{g}$ is referred to as the metric induced by $\iota$, where $\iota^\ast$ is the pullback. 
If $(M, g)$ is a Riemannian submanifold of $(\tilde{M}, \tilde{g})$, then for every $p \in M$ and $v, w \in T_p M$, the 
induced metric is defined as

\begin{equation} \label{eqn:induced_metric}
  g_p(v, w) = \tilde{g}_p(d \iota_p(v), d \iota_p(w)).
\end{equation}

This statement forms the guiding principle that allows us to compute metrics induced by differentiable function transformations.
We leverage this idea to define the metric induced by an autoencoder.

\subsection{Covariant Differentiation \& Parallel Transport}
Given a definition for the inner product on the manifold, we can introduce the notion of covariant differentiation; a 
means of specifying derivatives along tangent vectors. For a vector $w \in T_p M$ and a vector field $v \in TM$, we 
define the covariant derivative as

\begin{equation} \label{eqn:covariant_derivative}
  \nabla_v w = v^j \nabla_{\partial_j} w^i \partial_i = (v^j w^k_{;j}) \partial_k = \left( v^j w^k_{,j} + \Gamma^k_{ij} v^j w^i \right) \partial_k,
\end{equation}

where; for brevity, we introduce the notation $w^k_{,j}, w^k_{;j}$ to denote the partial, and covariant derivative respectively --
each taken with respect to the $j$th coordinate vector field. The expression containing the summation in Eq.~\ref{eqn:covariant_derivative}
provides an intuitive explanation for the concept of covariant differentiation: the first term represents the standard
directional derivative; and the second term is responsible for quantifying the 'twisting' of the coordinate system.
The term $\Gamma^k_{ij}$ is referred to as the Christoffel symbol of the second kind, or the affine connection; providing
a measure of how the basis changes over the manifold. We define the Christoffel symbols as

\begin{equation}
  \Gamma^k_{ij} = \partial_j(\partial_i) \cdot \partial^k = \tfrac{1}{2} g^{km} \left(
    g_{mi,j} + g_{mj,i} - g_{ij,m}
  \right),
\end{equation}

where $\partial^k = g^{km} \partial_{m}$ is the index-raised coordinate vector field, and $g^{km}$ is the inverse of the
metric tensor $g_{km}$. In the absence of covariant differentiation, transporting a vector $w \in T_p M$ along the
manifold does not guarantee it remains parallel to itself in the tangent planes along the path. Our definition of the
covariant derivative provides a sufficient condition to ensure the parallelism is conserved along a trajectory. Mathematically
we express this as

\begin{equation} \label{eqn:geodesic_equation}
  \nabla_v w =  v^j w^k_{,j} + \Gamma^k_{ij} v^j w^i = 0.
\end{equation}

The ability to preserve parallelism in the tangent plane along a trajectory on the manifold provides us with the ability
to define curves of interest, namely geodesics.

\subsection{Geodesics}
A geodesic on the manifold generalises the notion of a straight line between points. This geodesic curve $\gamma(\lambda)$ 
is locally length-minimising in the sense that it constitutes a solution of the Euler-Lagrange equations: it describes the
motion of a rigid body devoid of acceleration. Leveraging our definition of covariant differentiation, we can describe a 
geodesic as a curve $\gamma(\lambda)$ whose tangent vectors remain parallel to themselves as they are transported along
the curve, yielding the geodesic equation

\begin{equation}
  \nabla_{\dot{\gamma}} \dot{\gamma} = \ddot{\gamma}^k + \Gamma^k_{ij} \dot{\gamma}^i \dot{\gamma}^j = 0,
\end{equation}

where derivatives $\ddot{\gamma}, \dot{\gamma}$ are taken with respect to the affine parameter $\lambda$. This geodesic
equation ensures the speed $\langle \dot{\gamma}, \dot{\gamma} \rangle^{0.5}_g$ is constant along the trajectory, allowing
us to classify parameterisations $\lambda \in [0, 1]$ as unit-speed geodesics. Conversely, the geodesic can be parameterised
with respect to length to yield a unit-distance geodesic.

A geodesic $\gamma$ is said to be minimising \textit{iff} there exist no shorter, valid geodesics with the same endpoints.
The length of this minimising geodesic provides a notion of geodesic distance on the manifold. While we can validate that
a curve is a geodesic, there is no principled way to demonstrate that the length-minimising curve has been obtained. In
this work we address two key points: providing principled means to obtain valid geodesics on arbitrary manifolds, such as
those induced by machine learning models; and computing geodesic distance fields directly.

\subsection{Computational Framework}
In this paper we provide a computational framework for operating on differentiable manifolds. We leverage the automatic
differentiation cababilities of \textit{jax}~\citep{jax2018github} to compute the metric induced by an arbitrary, 
smooth immersion as described in Eq.~\ref{eqn:induced_metric}, as well as all derivative properties. We provide access 
to our open source library on Github. \footnote{All code is available on GitHub: \url{https://github.com/removed_for_anonymity}}

\section{Computing Geodesics} \label{sec:valid_geodesics}

Computing valid geodesics on differentiable manifolds has applications in many domains, not least computer 
vision~\citep{peyre2009GeodesicMethodsComputer}, general relativity~\citep{carroll2004SpacetimeGeometryIntroduction}, and
machine learning~\citep{bronstein2021GeometricDeepLearning}. In this section we introduce a principled numerical approach 
to computing valid geodesics on differentiable manifolds. We first define the initial value problem and demonstrate a numerical 
scheme which ensures energy conservation along the resultant geodesic. Next, we address the important problem of obtaining
geodesics connecting points on the manifold; this is presented as a boundary value problem for which we outline numerical
schemes which allow us to obtain valid geodesics. We then outline and discuss an approach for obtaining length-minimising
curves adopted by the machine learning community. Finally, we demonstrate the application of these methods on a manifold 
induced by an autoencoder, comparing the ability of these approaches to produce valid geodesics.

\subsection{On the Straight and Narrow: Initial Value Problem} \label{sec:initial_value_problem}
We first consider the initial value problem, defining unit-speed geodesics $\gamma: [0, 1] \rightarrow M$. Given a 
point $p \in M$, and an initial velocity vector $v \in T_p M$, we wish to compute the unique geodesic $\gamma(\lambda)$
such that $\gamma(0) = p, \dot{\gamma}(0) = v$. We express the initial value problem through the exponential map

\begin{equation} \label{eqn:initial_value_problem}
  \gamma(\lambda = 1) = \exp_p(v)
\end{equation}

The geodesic equation, as defined in Eq.~\ref{eqn:geodesic_equation}, is a second-order ordinary differential equation 
expressed in Lagrangian coordinates. As the geodesic equation is a solution to the Euler-Lagrange equations, this admits
the Hamiltonian

\begin{equation} \label{eqn:hamiltonian}
  H(q, p) = \frac{1}{2} g^{ij} p_i p_j,
\end{equation}

where $g^{ij}$ is the inverse metric tensor, and $p_i$ denotes the covariant conjugate momenta. This Hamiltonian formulation gives
rise to an alternative definition of the geodesic equations

\begin{equation} \label{eqn:hamiltonian_geodesic_equation}
  \dot{q}^i = \frac{\partial H}{\partial p_i} = g^{ij} p_j
\qquad
  \dot{p}_i = -\frac{\partial H}{\partial q^i} = - \frac{1}{2} g^{jk}_{,i} p_j p_k,
\end{equation}

which ensures energy conservation along the geodesics, as demontrated by

\begin{equation} \label{eqn:hamiltonian_energy_conservation}
  \frac{dH}{dt} = \frac{\partial H}{\partial q^k} \dot{q}^k + \frac{\partial H}{\partial p_k} \dot{p}_k = - \dot{p}_k \dot{q}^k + \dot{q}^k \dot{p}_k = 0.
\end{equation}

In computing the exponential map $\exp_p(v)$, we wish to leverage an integration scheme which conserves the Hamiltonian
along the resulting trajectory; a task best suited for symplectic integration. While classical integration schemes are
often subject to energy dissipation, symplectic integration methods are designed in such a manner as to provide an error 
bound for the energy dissipation; an important characteristic for energy-conserving systems. Conventional symplectic
integration schemes~\citep{2006GeometricNumericalIntegration}, such as the likes of symplectic Euler, velocity Verlet, 
rely on a separable Hamiltonian -- a property not exhibited by the geodesic equation.

We employ a symplectic integration scheme for nonseparable Hamiltonians proposed by~\citet{tao2016ExplicitSymplecticApproximation}.
The Hamiltonian is augmented in an extended phase space with symplectic 2-form $dq \wedge dp + dx \wedge dy$

\begin{equation} \label{eqn:augmented_hamiltonian}
  \bar{H}(q, p, x, y) \coloneqq H_A + H_B + \omega H_C,
\end{equation}

where $H_A = H(q, y), H_B = H(x, p)$ represent copies of the original system with mixed positions and momenta;
$H_C = \nicefrac{1}{2} \lVert q - x \rVert^2_2 + \nicefrac{1}{2} \lVert p - y \rVert^2_2$ is an artifical restraint for
the energy conservation; and $\omega$ is a constant which determines the strength of the relationship between the two
copies.

Strang-splitting~\citep{strang1968ConstructionComparisonDifference, mclachlan2002SplittingMethods} is employed to generate
a flow-map for a second-order integrator

\begin{equation} \label{eqn:flow_map_o2}
  \phi^\delta_2 \coloneqq 
  \phi^{\nicefrac{\delta}{2}}_{H_A} 
\circ 
  \phi^{\nicefrac{\delta}{2}}_{H_B}
\circ
  \phi^{\delta}_{\omega H_C}
\circ
  \phi^{\nicefrac{\delta}{2}}_{H_B}
\circ
  \phi^{\nicefrac{\delta}{2}}_{H_A},
\end{equation}

where individual flow-map components constitute the integration step. This integration scheme can be extended to
arbitrary higher-order schemes through use of the Yoshida triple-jump~\citep{yoshida1990ConstructionHigherOrder}

\begin{equation}
  \phi^\delta_l \coloneqq 
  \phi^{\psi_l \delta}_{l - 2}
\circ
  \phi^{(1 - 2 \psi_l) \delta}_{l - 2}
\circ
  \phi^{\psi_l \delta}_{l - 2},
\qquad \text{where} \quad
  \psi_l = \frac{1}{2 - 2^{\nicefrac{1}{(l + 1)}}},
\end{equation}

and $l$ is the desired order of integration. For the purposes of this work, we employ a fourth-order symplectic
integration scheme, $l = 4$.

\subsection{Connecting the Dots: Boundary Value Problem} \label{sec:boundary_value_problem}
Given points $p, q \in M$, we wish to find the geodesic $\gamma: [0, 1] \rightarrow M$ which connects these points. Using
our definition of the exponential map, we can express this as $q = \exp_p(v)$, where we wish to find the initial tangent
vector $v \in T_p M$. This formulation results in a two-point boundary value problem for which standard shooting methods
can be applied~\citep{keller1976NumericalSolutionTwo}. We define the residual

\begin{equation} \label{eqn:bvp_residual}
  r(v) = \exp_p(v) - q = 0,
\end{equation}

where the exponential map $\exp_p(v)$ is computed using the symplectic integration scheme described in §\ref{sec:initial_value_problem}.
Roots of the residual $r(v)$ denote solutions to the boundary value problem, expressable using the logarithmic map
$v = \log_p(q)$; a natural inverse to the exponential map~\citep{carmo2018DifferentialGeometryCurves}. We employ the 
standard Newton-Raphson root-finding method to obtain solutions to the boundary value problem and find valid geodesics
connecting two points. We note that this shooting method finds locally length-minimising geodesics, not necessarily the
shortest geodesics.

\subsection{Length-Minimising Geodesics} \label{sec:length_minimising_geodesics}
Geodesics are defined as locally shortest paths on the manifold -- solutions to the Euler-Lagrange equations. In many
circumstances there are multiple geodesics connecting two points, each with different lengths. We define the length of 
a geodesic as

\begin{equation} \label{eqn:geodesic_length}
  L(\gamma) = \int_0^1 \langle \dot{\gamma}(t), \dot{\gamma}(t) \rangle^{\nicefrac{1}{2}}_{\gamma(t)} dt.
\end{equation}

The geodesic distance $d_g(p, q)$ between two points $p, q \in M$ is defined as infimum of the length of all valid
geodesics which connect the two points; that is 

\begin{equation} \label{eqn:geodesic_distance}
  d_g(p, q) = \inf_\gamma \left\{ L(\gamma) : \gamma(0) = p, \gamma(1) = q \right\}.
\end{equation}

Conversely, this admits that the shortest curve between two points is itself a valid geodesic. Work by~\citet{chen2018MetricsDeepGenerative} 
leverages the notion that the shortest curve connecting two points on the manifold yields a valid geodesic, admitting the 
geodesic distance $d_g(p, q)$. In their proposed methodology a neural network $\gamma_\theta: [0, 1] \rightarrow M$, 
constrained with the boundary values $\gamma_\theta(0) = p, \gamma_\theta(1) = q$, parameterises the target geodesic.
Parameters $\theta$ are optimised in accordance with the objective of minimising the length of the resultant curve, as 
per Eq.~\ref{eqn:geodesic_length}. This proposed approach has been adopted by the machine learning community to compute 
distances and produce geodesic paths on latent manifolds~\citep{tennenholtz2022UncertaintyEstimationUsing, arvanitidis2020GeometricallyEnrichedLatent}. 

Obtaining valid geodesics, and thus the geodesic distance, is reliant on the solution to a global optimisation problem.
In the event that the resultant curve is not a geodesic, there must exist a valid geodesic with shorter length. That is
to say

\begin{equation} \label{eqn:distance_inequality}
  d_g(p, q) < L(\gamma_\theta) \quad \text{when} \quad \theta \neq \theta^{*},
\end{equation}

where $\theta^*$ denotes the globally optimum parameters. There is also no guarantee that the network architecture used
to parameterise $\gamma_\theta$ even has the capacity to represent the true geodesic. Although the approach is well-justified, 
we find that the model is prone to getting stuck in local minima and that converged solutions do not produce valid geodesics.
We examine the ability of this approach to produce valid geodesics in §\ref{sec:latent_interpolation} and compare results
to the principled numerical integration approach.

\subsection{Interpolation on an Autoencoder manifold.} \label{sec:latent_interpolation}
We compare the aforementioned approaches by computing paths on a manifold defined by an autoencoder 
$\eta_\theta = \eta^e_\theta \circ \eta^d_\theta$, where $\eta^e_\theta: \mathbb{R}^n \rightarrow M$ represents the
encoder, and $\eta^d_\theta: M \rightarrow \mathbb{R}^n$ the decoder. An autoencoder with latent dimensionality
$z \in M \subset \mathbb{R}^2$ is pre-trained on the MNIST dataset~\citep{deng2012mnist}, with the resulting latent space
representing intrinsic coordinates on the manifold. Using our definition of the induced metric per Eq.~\ref{eqn:induced_metric}, 
we define the metric induced by the decoder ~\citep{magri2022InterpretabilityProperLatent, chen2018MetricsDeepGenerative} as

\begin{equation}
  g_{ij}(z) = \frac{\partial \eta^d_\theta(z^k)}{\partial z^i} \frac{\partial \eta^d_\theta(z^k)}{\partial z^j}.
\end{equation}

In order to provide a visualisation of the  manifold, we compute the magnification factor, \textit{MF}, across the domain. 
The magnification factor denotes the rate of change of infintesimal volume, mapping a point on the manifold to Euclidean 
space; precisely $\textit{MF} = \sqrt{\det{g_{ij}}}$.

\begin{figure}[ht]
\centering
\begin{subfigure}[ht]{\linewidth}
  \centering
  \includegraphics[width=\linewidth]{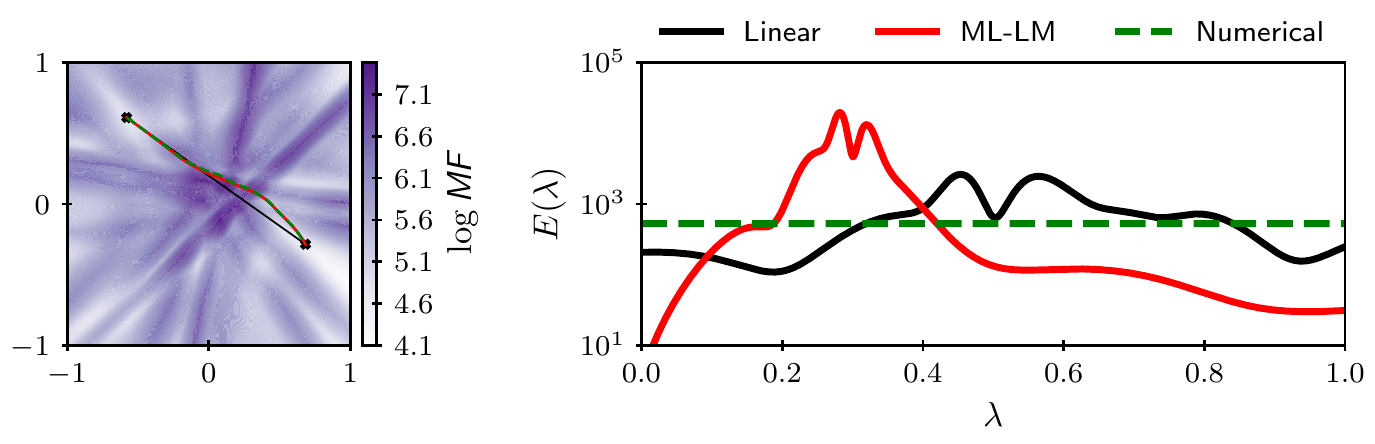}
  \caption{
    Latent trajectories $\gamma(\lambda)$ for the three interpolation schemes are shown against a background of magnification
    factor, \textit{MF} on the left; and energy, $E = \langle \dot{\gamma}, \dot{\gamma} \rangle_g$ along each trajectory
    is shown on the right. We observe constant energy for the numerical trajectory, a result of solving the geodesic equation
    directly on the manifold.
  }
  \label{fig:latent_trajectories}
\end{subfigure}
\par\bigskip 
\begin{subfigure}[ht]{\linewidth}
  \centering
  \includegraphics[width=\linewidth]{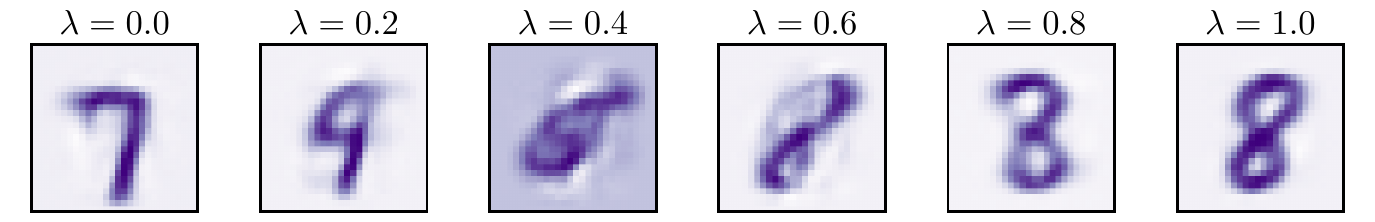}
  \caption{
    Decoder realisations $\eta^d_{\theta}(\gamma(\lambda))$ of samples drawn from numerical geodesic trajectory. By conserving
    energy along the interpolation trajectory we are able to obtain smooth transitions between points on the manifold.
  }
  \label{fig:trajectory_realisations}
\end{subfigure}
  \caption{
    Interpolation on the manifold induced by an autoencoder trained on the MNIST dataset.
  }
  \label{fig:latent_interpolation_results}
\end{figure}

We consider trajectories $\gamma(\lambda)$ between points $z_1, z_2 \in M$ on the resulting manifold. Results in 
Figure~\ref{fig:latent_trajectories} demontrate three independant trajectories between these points on the manifold,
corresponding to: linear interpolation; the length-minimising (ML-LM) curve as described in §\ref{sec:length_minimising_geodesics}; 
and a valid geodesic produced by integration of the geodesic equation, as described in §\ref{sec:boundary_value_problem}.
In each case, we obtain a curve $\gamma: [0, 1] \rightarrow M$. A time-step of $\Delta t = 10^{-3}$ is employed, yielding
a trajectory with $1000$ time-steps. The network parameterising the length-minimising curve is obtained using an ensemble
of $30$ networks, each trained with the \textit{adam} optimiser~\citep{kingma2017AdamMethodStochastic}. For the numerical 
integration, a fourth-order symplectic integration scheme is used with $\omega = 10^{-2}$. Further details of network 
architecture and training details can be found in the supplementary material.

Trajectories for the length-minimising curve and the valid geodesic are noticably similar with regard to the resulting
coordinates, however the energy along each trajectory provides more insight. We observe that the trajectories produced
by linear interpolation and the length-minimising curve approach are subject to large degrees of energetic deviation,
violating the Hamiltonian-preserving properties of the geodesic equation. These energy fluctuations present in the ML-LM
approach invalidate the ability to compute distance in a meaningful way, a fundamental flaw of the existing approach. The
trajectory obtained from numerical integration yields a valid geodesic, exhibiting energy conservation along the path;
this makes it much more effective for latent interpolation as well as distance computation.

Samples can be drawn along the resulting trajectory to provide a form of latent interpolation. Realisations shown in
Figure~\ref{fig:trajectory_realisations} depict samples drawn along the valid numerical trajectory, drawing samples
$\eta_\theta(\gamma(\lambda)), \lambda \in \{0.0, 0.2, 0.4, 0.6, 0.8, 1.0\}$. Interpolation along an energy-conserving
curve yields a smoother transition when compared with the alternative approaches. This has numerous applications for
latent space interpolation, such as that used in generative modeling.

\section{Distance Fields: Solutions to the Eikonal Equation} \label{sec:eikonal_equation}

Obtaining length-minimising geodesics between pairwise points is challenging, as dicussed in §\ref{sec:valid_geodesics};
instead, we introduce an alternative approach to computing geodesic distances. Given a point $p \in M$, we consider a 
function $\phi: p \mapsto d_g(p, q)$ which computes the geodesic distance to an arbitrary point $q \in M$. The resultant
distance function is a solution to the Eikonal equation

\begin{equation} \label{eqn:eikonal}
  \lVert \nabla \phi \rVert = 1 \qquad \text{s.t.} \quad \restr{\phi}{p} = 0.
\end{equation}

We generalise the notion of the Eikonal equation to the manifold, leveraging our definition of the inner product and
using the exterior derivative. Upon manipulation, this yields

\begin{equation} \label{eqn:manifold_eikonal}
  \langle grad \; \phi, grad \; \phi \rangle_{g} = g_{ij} \phi^{;i} \phi^{;j} = 1,
\end{equation}

where $\phi^{;i} = g^{ij} \phi_{;j}$ denotes the index-raised covariant derivative. For a distance function $\phi$,
the gradient of the distance field provides the geodesic flow, $\nabla \phi$. This flow satisfies the geodesic equation

\begin{equation} \label{eqn:geodesic_flow}
  \nabla_{\nabla \phi} \nabla \phi = 0,
\end{equation}

where integral paths constitute unit-distance geodesics travelling orthogonal to the level-sets imposed by the distance
field. As a by-product of computing the distance function $\phi$, we are able to compute the length-minimising geodesics. 
We refer the reader to \citet[see][Thm.~6.31,~6.32]{lee2018IntroductionRiemannianManifolds} for a more rigorous overview
on geodesic flows, and solutions to the Eikonal equation on the manifold.

While solutions to the Eikonal equation on the manifold have been considered previously, these methods leverage a series
of approximations and discretisations to obtain solutions. Most notable are heat kernel methods which rely on numerical
approximations to the Laplace-Beltrami operator, the dirac-delta distribution, and Varadhan's 
formula~\citep{crane2013GeodesicsHeatNew, varadhan1967BehaviorFundamentalSolution, gropp2020ImplicitGeometricRegularization}.
We propose an approach for working with continuous function representations, relying on no such approximations.

We parameterise the distance field by a neural network $\phi_\theta: M \rightarrow \mathbb{R}_{\geq 0}$, architecturally
constrained in such a manner as to ensure $\phi_\theta(p) = 0$. The solution is posed as an optimisation problem

\begin{equation} \label{eqn:optimisation}
  \theta^* = \argmin_{\theta} \mathcal{L}(\epsilon_\phi, \epsilon_{\nabla \phi}) 
    \quad \text{where} 
    \; 
      \epsilon_\phi = g_{ij} \phi^{;i}_{\theta} \phi^{;j}_{\theta} - 1,
    \;
      \epsilon_{\nabla \phi} = \langle  \nabla_{\nabla \phi_\theta} \nabla \phi_\theta, \nabla_{\nabla \phi_\theta} \nabla \phi_\theta \rangle_{q}
\end{equation}

Naturally, the solution to the Eikonal equation is more sensitive in regions of high curvature; a consequence of induced
geodesic deviation. We introduce a novel sampling and loss-scaling methodology to ensure we are able to capture the solution
appropriately in regions of high curvature.

\subsection{Introducing the Curvature}
The Riemann curvature tensor $R^{l}_{ijk}$ provides a means to represent the curvature on the manifold. Mathematically,
this curvature denotes the failure of the second covariant derivatives to commute, constituting the tidal force experienced
by a rigid body moving along a geodesic. A Riemannian manifold has zero curvature in regions exhbiting local isometry to
Euclidean space. In the interest of obtaining a scalar value for the curvature, we introduce the following progression

\begin{equation} \label{eqn:curvature_progression}
  R^l_{ijk} = \Gamma^l_{ik,j}- \Gamma^l_{ij,k} + \Gamma^l_{jm} \Gamma^m_{ik} - \Gamma^l_{km} \Gamma^m_{ij}
  \quad \rightarrow \quad
  R_{ij} = R^{m}_{\phantom{m}imj}
  \quad \rightarrow \quad
  R = g^{ij} R_{ij}
\end{equation}

where $R^l_{ijk}$ denotes the Riemann curvature tensor; $R_{ij}$ the Ricci curvature tensor, obtained by contracting 
over the first and third indices; and $R$ is the Ricci scalar, the trace of the Ricci tensor with respect to the metric. 
The Ricci scalar is a local invariant on the manifold and assigns a single real number to the curvature at a particular 
point. We note that the demonstrated progression relies solely on the definition of the metric tensor $g_{ij}$.

As a result of geodesic deviation exhibited in regions of high curvature, the geodesic flow $\nabla \phi$, and 
consequently the distance field $\phi_\theta$, are more challenging to characterise. We leverage information about the
curvature to provide a training mechanism which takes this into account.

\subsection{Leveraging the Curvature}

We propose a novel sampling and scaling strategy for the loss, utilising information about the local scalar curvature of
the manifold. Employing the Metropolis-Hastings \citep{metropolis1953equation, hastings1970monte} sampling algorithm, 
we draw samples in regions of the manifold which exhibit high-degrees of local scalar curvature. To ensure a representative
distribution, we sample multiple chains and discard initial samples to ensure convergence of the Markov chain. A kernel
density esimate is obtained for the resulting points to obtain a probability density function to allow efficient sampling
in the training process.

We propose a loss function which scales the residuals based on the degree of local curvature

\begin{equation} \label{eqn:curvature_scaled_loss}
  \mathcal{L} = \frac{1}{\lvert \mathcal{X} \rvert} \sum_{x \in \mathcal{X}}
    \psi(x) \left[ \epsilon_\phi(x)^2 + \lambda \epsilon_{\nabla \phi}(x) \right]
  \qquad \text{where} \quad \psi(x; \alpha) = 1 + \alpha \log{\left( 1 + R(x) \right)},
\end{equation}

where $x \in \mathcal{X}$ are training samples; $\lambda$ is a weighting term for the residual losses; and $\alpha$ is
a scaling factor used to determine the relative importance of the local curvature. For the sake of this work, we take
$\lambda = 10^{-3}, \alpha=10^{-1}$, empirically determined to provide satisfactory convergence. Training samples 
$\mathcal{X}$ are drawn from a distribution $\mathcal{D}$ which is the average of a uniform distribution over the domain,
and the probability density function generated from the curvature samples. This proposed loss function is used as the
objective function in the optimisation problem posed in Eq.~\ref{eqn:optimisation}.

In order to provide an interpretable example, we demonstrate results for a function $f: \mathbb{R}^2 \rightarrow \mathbb{R}^3$, 
where intrinsic coordinates are interpretable in a two-dimensional space. We choose a complex manifold with a multitude
of local extrema, based on the \textit{peaks} function~\citep{PeaksFunctionMATLAB}, namely

\begin{equation}
  f: (x, y) \mapsto 3 (1 - x)^2 e^{-x^2 - (y + 1)^2} - 10 \left(\tfrac{x}{5} - x^3 - y^5 \right) e^{-x^2-y^2} - \tfrac{1}{3} e^{-(x + 1)^2 - y^2}.
\end{equation}

\begin{figure}[ht]
     \centering
     \begin{subfigure}[ht]{0.32\textwidth}
         \centering
         \includegraphics[width=\textwidth]{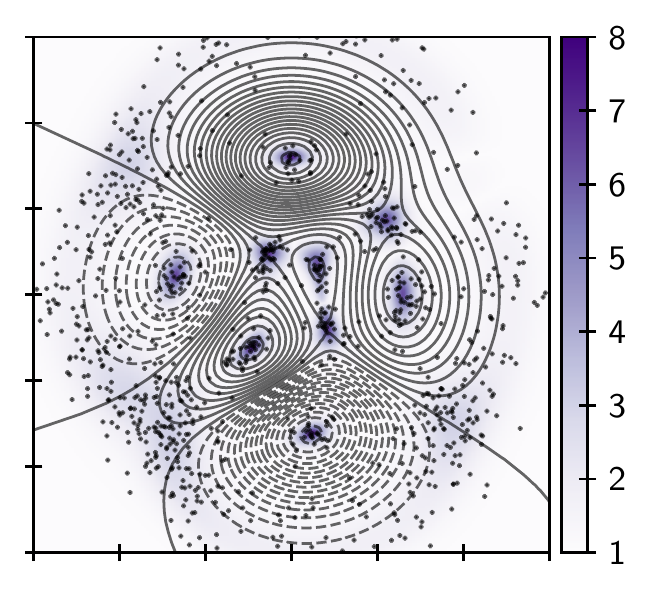}
         \caption{Sampling and Scaling}
         \label{fig:sampling_and_scaling}
     \end{subfigure}
     \hfill
     \begin{subfigure}[ht]{0.32\textwidth}
         \centering
         \includegraphics[width=\textwidth]{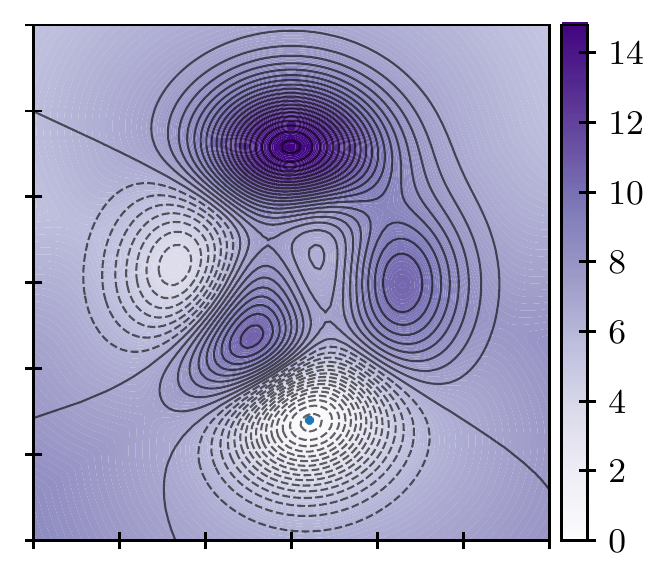}
         \caption{Distance field, $\phi_\theta$}
         \label{fig:geodesic_distance}
     \end{subfigure}
     \hfill
     \begin{subfigure}[ht]{0.32\textwidth}
         \centering
         \includegraphics[width=\textwidth]{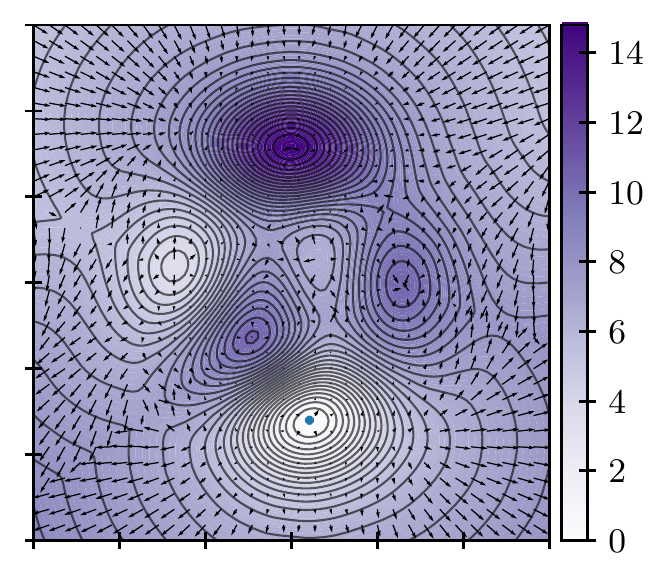}
         \caption{Geodesic flow, $\nabla \phi_\theta$}
         \label{fig:geodesic_flow}
     \end{subfigure}
        \caption{
          Solutions to the Eikonal equation: 
            (\subref{fig:sampling_and_scaling}) demonstrates the curvature scaling field, $\psi(x; \alpha=1)$, and 
                corresponding samples drawn from the scalar curvature distribution; 
            (\subref{fig:geodesic_distance}) shows the geodesic distance predicted by the network, $\phi_\theta$; and
            (\subref{fig:geodesic_flow}) shows the resulting geodesic flow, $\nabla \phi$.
        }
        \label{fig:three graphs}
\end{figure}

A demonstration of the manifold, as well as the loss scaling and sampling strategy can be seen in Figure~\ref{fig:sampling_and_scaling}. 
We observe that regions of high scalar curvature are subjected to higher degrees of penalisation in the loss term, and 
additional samples for training are drawn from the curvature distribution. This combination of scaling and sampling is 
effective in overcoming challenges to obtaining a solution to the Eikonal equation.

Results in Figure~\ref{fig:geodesic_distance} demonstrate the predicted geodesic distance on the manifold from a point
$p$. We observe that the distance field conforms to the manifold, as represented by the level-set contours; this
is a consequence of re-formulating the Eikonal equation to account for the imposed metric tensor. The level-sets of the
resultant distance field are shown in Figure~\ref{eqn:geodesic_flow}, with arrows showing the geodesic flow $\nabla \phi_\theta$
on the manifold. As expected, the geodesics flow orthogonal to the level sets imposed by the distance field $\phi_\theta$
and provide geodesic paths.

\newpage

\section{Conclusions} \label{sec:conclusion}

In this work we first introduce a numerical methodology for computing geodesics on differentiable manifolds. An autoencoder
is pre-trained on the MNIST dataset, inducing a Riemannian metric which allows us to operate in a principled manner on
the resulting latent space. We consider the task of interpolation in the latent space and compare trajectories obtained
by: linear interpolation; a machine learning approach for obtaining length-minimising curves; and our numerical method.
Conservation of energy along the latent trajectory is a core principle for valid geodesics. We demonstrate that only the
numerical approach was able to achieve valid geodesics, satisfying the geodesic equation, and demonstrating energy
conservation. The ability to produce valid geodesics allows for smooth interpolation, imperative for applications in
generative modelling, and reduced-order modelling.

We then address the problem of computing geodesic distance, providing a means to obtain a model-based parameterisation
of distance fields and geodesic flows on differentiable manifolds. Our proposed methodology relies on a manifold-informed
extension to the Eikonal equation, allowing us to obtain a continuous function representation, without unnecessary
assumptions or approximations. Key to this is a novel sampling and scaling training mechanism which leverages information
about the curvature of the manifold to stabilise training in regions subject to significant geodesic deviation. We demonstrate
the ability to produce distance fields which conform to the manifold, a consequence of imposing knowledge of the metric
throughout the training process.

\acksection{
  D. Kelshaw. and L. Magri. acknowledge support from the UK EPSRC and thank Marika Taylor for fruitful discussions.
  L. Magri gratefully acknowledges financial support from the ERC Starting Grant PhyCo 949388.
}

\bibliography{references}

\section*{Supplementary Material}

\subsection*{Training Details: Latent Interpolation}

For the task of latent interpolation, we train a standard autoencoder on the
MNIST dataset. The network $\eta_\theta$, tasked with learning the identity function
$\eta_\theta: x \mapsto x$, consists of an encoder $\eta^e_\theta$ and decoder 
$\eta^d_\theta$. A precise definition is given by

\begin{equation*}
\begin{gathered}
	\eta_\theta = \eta^e_\theta \circ \eta^d_\theta, \quad \text{where} \quad
		\eta^e_\theta = \sigma(\eta^{e1}_\theta) \circ \sigma(\eta^{e2}_\theta),
	\quad 
		\eta^d_\theta = \sigma(\eta^{d1}_\theta) \circ \eta^{d2}_\theta
\\[15pt]
\begin{aligned}
	\eta^{e1}_\theta &: \mathbb{R}^{784} \rightarrow \mathbb{R}^{32} \\
	\eta^{e2}_\theta &: \mathbb{R}^{32} \rightarrow \mathbb{R}^{2}
\end{aligned}
	\qquad \qquad
\begin{aligned}
	\eta^{d1}_\theta &: \mathbb{R}^{2} \rightarrow \mathbb{R}^{32} \\
	\eta^{d2}_\theta &: \mathbb{R}^{32} \rightarrow \mathbb{R}^{784}
\end{aligned}
\end{gathered}
\end{equation*}

All intermediate layers $\eta^{(\cdot)}_\theta$ are linear transformations, using
activations $\sigma: x \mapsto \textit{tanh}(x)$ to induce nonlinearities. The
\textit{adam} optimiser is used for training, employing a learning rate of $3 \times 10^{-4}$.

Data $x \in \mathcal{X}$ is standardised to the range $x \in [0, 1]$ before being
fed to the network. We employ the standard data split using $50000$ samples for
training and $10000$ for validation. Optimal parameters $\theta^\ast$ are chosen
to be those which minimise the reconstuction error on the validation set.

\subsection*{Training Details: Eikonal Equation}

The distance function is parameterised by a neural network $\phi_\theta: M \rightarrow \mathbb{R}_{\geq 0}$
constrained to provide strictly positive outputs with $d_g(q, q) = 0$. This is
achieved by transforming the outputs of a network

\begin{equation*}
	\phi_\theta: (p, q) \mapsto \left\lvert \tilde{\phi}_\theta(p) - \tilde{\phi}_\theta(q) \right\rvert,
\end{equation*}

where the unconstrained network $\tilde{\phi}_\theta$ is defined by a sequence of transformations

\begin{equation*}
\begin{gathered}
	\tilde{\phi}_\theta = 
		\sigma \left( \tilde{\phi}^{(1)}_\theta \right)
	\circ 
		\sigma \left( \tilde{\phi}^{(2)}_\theta \right) 
	\circ
		\sigma \left( \tilde{\phi}^{(3)}_\theta \right)
	\circ
		\tilde{\phi}^{(4)}_\theta.
\\[15pt]
\begin{aligned}
	\tilde{\phi}^{(1)}_\theta &: M \rightarrow \mathbb{R}^{64}
\\
	\tilde{\phi}^{(2)}_\theta &: \mathbb{R}^{64} \rightarrow \mathbb{R}^{64}
\end{aligned}
	\qquad
\begin{aligned}
	\tilde{\phi}^{(3)}_\theta &: \mathbb{R}^{64} \rightarrow \mathbb{R}^{64}
\\
	\tilde{\phi}^{(4)}_\theta &: \mathbb{R}^{64} \rightarrow \mathbb{R}^{1}
\end{aligned}
\end{gathered}
\end{equation*}

Each transformation $\tilde{\phi}^{(\cdot)}_\theta$ is a linear layer, and we
choose an activation $\sigma: x \mapsto \textit{tanh}(x)$ to induce nonlinearities.
We consider the domain $p \in [-3, 3] \subset \mathbb{R}^{2}$ and standardise inputs
to the network to the range $[-1, 1]$. Training is conducted using a set of random 
samples drawn from the domain. A total of $20000$ samples are drawn from the domain
at each epoch: half drawn from a uniform distribution; the remaining half drawn 
from the distribution defined by the curvature. We employ the \textit{adam} optimiser
with a learning rate of $3 \times 10^{-4}$ and train until we observe convergence.

\end{document}